\documentclass[conference]{IEEEtran}
\IEEEoverridecommandlockouts
\usepackage{cite}
\usepackage{amsmath,amssymb,amsfonts}
\usepackage{algorithmic}
\usepackage{graphicx}
\usepackage{textcomp}
\usepackage{xcolor}
\usepackage{fancyhdr}
\usepackage{url}
\usepackage{graphicx}
\usepackage{subcaption}
\usepackage{stfloats}  
\usepackage{placeins}
\usepackage{float}
\usepackage{adjustbox}
\usepackage{booktabs}
\usepackage{afterpage}
\usepackage[hidelinks]{hyperref} 
\def\BibTeX{{\rm B\kern-.05em{\sc i\kern-.025em b}\kern-.08em
    T\kern-.1667em\lower.7ex\hbox{E}\kern-.125emX}}
    
\begin{document}

\title{When Active Learning Falls Short: An Empirical Study on Chemical Reaction Extraction\\

}

\author{\IEEEauthorblockN{1\textsuperscript{st} Simin Yu}
\IEEEauthorblockA{\textit{Data and Knowledge Engineering} \\
\textit{Otto-von-Guericke University}\\
Magdeburg, Germany \\
simin.yu@st.ovgu.de}
\and
\IEEEauthorblockN{2\textsuperscript{nd} Sufia Fathima}
\IEEEauthorblockA{\textit{Data and Knowledge Engineering} \\
\textit{Otto-von-Guericke University}\\
Magdeburg, Germany\\
sufia@ovgu.de}

}

\maketitle
\thispagestyle{plain}
\pagestyle{plain}

\begin{abstract}
The rapid growth of chemical literature has generated vast amounts of unstructured data, where reaction information is particularly valuable for applications such as reaction predictions and drug design. However, the prohibitive cost of expert annotation has led to a scarcity of training data, severely hindering the performance of automatic reaction extraction. In this work, we conduct a systematic study of active learning for chemical reaction extraction. We integrate six uncertainty- and diversity-based strategies with pretrained transformer-CRF architectures, and evaluate them on product extraction and role labeling task. 
While several methods approach full-data performance with fewer labeled instances, learning curves are often non-monotonic and task-dependent. Our analysis shows that strong pretraining, structured CRF decoding, and label sparsity limit the stability of conventional active learning strategies. These findings provide practical insights for the effective use of active learning in chemical information extraction.
\end{abstract}


\section{Introduction}
With the exponential growth of chemical literature, it has become increasingly difficult for researchers to manually extract useful information from past publications. Knowledge of existing chemical reactions serves as a crucial foundation for chemists in conducting benchmark experiments, reaction predictions\cite{coley2019graph}, and computer-aided drug design\cite{sadybekov2023computational}. Despite the abundance of chemical reactions documented in journal articles, it remains challenging to convert them into structured formats. 
The difficulty arises because reaction extraction involves extracting both chemical entities and the relationships between them, which are reflected not necessarily in the same sentence or even the same paragraph\cite{guo2021automated}. Existing large-scale databases, such as Reaxys\cite{reaxys} and Scifinder\cite{scifinder}, heavily depend on manual extraction by human experts, which entails prohibitive costs, restrictive access and delayed updates\cite{lawson2014making}. 

As a result, attention has shifted toward developing effective automated methods for reaction extraction. Early approaches relied on manually designed rules and regular expressions to identify reaction components and their roles, leading to tools such as ChemicalTagger\cite{hawizy2011chemicaltagger} and OPSIN\cite{lowe2012extraction}. These rule-based systems were initially designed for patent data, which typically follows standardized formats and structured templates. OPSIN, partially built upon ChemicalTagger, assigned four roles to chemical entities, including product, reactant, solvent and catalyst. Nevertheless, these systems fail for multi-step reactions\cite{lowe2012extraction} and scale poorly to scientific articles, due to the complexity and variations in language use\cite{guo2021automated}. These limitations have motivated a shift from rule-based to data-driven approaches that directly learn from examples rather than depending on exhaustive rule design. 

Subsequent studies adopt traditional machine learning methods such as Conditional Random Fields(CRF) for chemical named entity recognition(NER), resulting in systems like ChemSpot\cite{rocktaschel2012chemspot} and tmChem \cite{leaman2015tmchem}.  Mallory et al. \cite{mallory2020extracting} applied a weakly supervised learning framework, Snorkel\cite{ratner2017snorkel}, to perform reaction extraction from biomedical literature abstracts. However, these approaches still fall short of producing fully structured reaction data, which includes chemical entities and their respective roles in the reactions.

Recent advances in Large Language Models (LLMs) have substantially improved performance on chemical Natural Language Processing(NLP) tasks, including NER\cite{gu2021domain} and relation extraction\cite{zhang2024fine}. Among these tasks, reaction extraction remains under-explored, despite its greater potential to generate informative and actionable data\cite{guo2021automated}. BERT-based models\cite{devlin2019bert} are typically pretrained on vast amounts of unlabeled corpora and subsequently fine-tuned for specific tasks. Following this paradigm, Guo et al.\cite{guo2021automated}  first proposed to decompose reaction extraction problem into a cascade of product extraction and reaction role labeling, corresponding to named entity recognition and conditioned relation extraction tasks in NLP, respectively. Their approach leverages ChemBERT, a transformer encoder pretrained on extensive chemical journal articles followed by a CRF decoder, and ChemRxnBERT, a variant further pretrained on reaction-relevant data. Their model achieved F1 scores of 76.2\% for product extraction and 78.7\% for role labeling, using only a few hundred annotated reactions. 

Despite these encouraging results, annotation of chemical reaction data remains time-consuming and expertise-demanding. While pretrained transformers reduce data requirements compared to earlier approaches, it is still an open challenge to further improve annotation efficiency. Active Learning(AL) offers a new perspective, by selecting informative or representative samples from large unlabeled corpora\cite{settles2009active} rather than randomly sampling, potentially improving the performance-to-annotation ratio. However, it remains unclear whether active learning can be effective when applied to pretrained transformer-CRF architectures in chemical reaction extraction. As pointed out by Guo et al.\cite{guo2021automated}, structured extraction of reaction data involves sequence tagging, severe class imbalance, and structured decoding constraints, all of which may affect acquisition strategies that are originally mainly designed for classical classification settings. 

In this work, we systematically investigate the behavior of uncertainty-based and diversity-based active learning strategies when integrated with ChemBERT and ChemRxnBERT. Our experiments demonstrate that active label acquisition and iterative fine-tuning with warm start can produce near-baseline results with reduced annotated data. However, none of the active learning strategy consistently outperform passive learning, and learning curves exhibit non-monotonic behavior. Our analysis reveals that heavily pretrained models, structured prediction with CRF decoding and tasks with extreme label sparsity may be incompatible with classic active learning strategies, providing insights into how active learning should be utilized in chemical reaction extraction. 

\section{Related Work}
Large publicly available datasets such as CoNLL-2003 \cite{sang2003introduction} and OntoNotes \cite{weischedel2011ontonotes} have played a crucial role in general-domain NER. However, due to annotation being expensive and requiring domain knowledge, specialized fields such as chemistry lack comparable resources \cite{jehangir2023survey}. This section reviews prior work addressing data scarcity through active learning, as well as research on chemical information extraction. 

\subsection{Active Learning for Deep Learning}
Active learning(AL) is a machine learning approach in which models actively choose the most informative samples for labeling rather than learning from randomly chosen data. Motivated by high annotation costs and limited labeled data\cite{shen2017deep}, AL has increasingly been integrated with deep learning, resulting in what is often referred to as Deep Active Learning (DeepAL) \cite{ren2021survey}. This integration leverages the strengths of both, as deep learning effectively extracts features from high-dimensional inputs, while active learning reduces labeling requirements\cite{ren2021survey}. 

Many DeepAL methods are based on uncertainty sampling, where models query samples for which predictions are least confident\cite{yang2016active}. Cost-Effective Active Learning, developed by Wang et al. \cite{wang2016cost}, combines Convolutional Neural Network (CNN) with uncertainty-based active acquisition. It includes least-confidence and margin sampling, and additionally used pseudo-labeling of high-confidence predictions based on entropy thresholds to expand the training set. Their approach reduced annotation needs by 36\%. Gal et al. \cite{gal2017deep} extended this idea through Deep Bayesian Active Learning, which employs Monte Carlo dropout to obtain more reliable uncertainty estimates for sample selection. Similarly, Shelmanov et al. \cite{shelmanov2021active} introduced a transformer-based active learning framework using Maximum Normalized Log-Probability, Variation Ratio, and Bayesian Active Learning by Disagreement(BALD). Their experiments demonstrate that high performance can be achieved for low-resource NLP tasks using only 16--20\% of labeled data.  

Diversity-based methods provide an alternative to uncertainty sampling, addressing its tendency to select redundant or highly similar queries \cite{ren2021survey}. The Core-set approach proposed by Sener and Savarese\cite{sener2017active} formulates active learning as choosing a representative subset that best cover the feature space distribution. A more recent study by Liu et al. \cite{liu2024utilizing} proposed CLUSTER, a diversity-based method for clinical NER that groups semantically similar sentences using sentence-BERT embeddings in combination with K-means clustering. Their framework also incorporates dynamic strategies, such as cluster least confidence and cluster N-best sequence entropy, that enabled a transition from diversity-based to uncertainty-based selection as learning progresses. 

Hybrid strategies attempt to strike a balance between uncertainty and diversity, ensuring that selected samples are simultaneously informative and representative of the data distribution \cite{ren2021survey}. Li et al. \cite{li2022ud_bbc} proposed such an approach within a BERT-BiLSTM-CRF architecture for named entity recognition. This hybrid strategy calculates uncertainty scores with least confidence and diversity scores with POS-tagged context similarity, thereby minimizing redundant annotations. Likewise, Liu et al. \cite{liu2024cybersecurity} combined normalized least-confidence sampling with self learning based on maximum confidence for cybersecurity NER under restricted labeled data. The approach achieved 97.5\% of full-data performance using only 30\% of labeled samples, resulting in an F1-score of 94.04\%. Although active learning has shown success in low-resource NLP scenarios, the use of AL strategies in domain-specific pretrained transformer-CRF architectures with high label imbalance has not yet been explored. 

\subsection{Chemical Information Extraction}
Chemical information extraction involves automatically identifying structured information about chemical entities, reactions, and their properties from unstructured scientific text, which enables applications such as drug discovery and reaction database construction \cite{swain2016chemdataextractor, lowe2012extraction}. This task is particularly challenging due to the complexity of chemical nomenclature, which includes systematic IUPAC names, chemical formulas, trivial names, and trade names that often appear interchangeably within documents \cite{krallinger2015chemdner}. 

Early systems such as OSCAR4 \cite{jessop2011oscar4} and ChemSpot \cite{rocktaschel2012chemspot} relied on dictionary matching and rule-based approaches along with CRF to identify various chemical terms \cite{khabsa2015towards, eltyeb2014chemical}. Although these approaches achieved moderate success, they depended heavily on manual feature engineering, which performed less effectively on journal articles compared to patents due to the varied writing style and less formulaic language \cite{krallinger2017information}. ChemicalTagger \cite{hawizy2011chemicaltagger}, a rule- and grammar-based parser, went beyond entity extraction by recognizing action phrases and relationships, but its reliance on handcrafted rules restricted scalability. The BiLSTM-CRF architecture \cite{lample2016neural}which established a strong foundation for sequence tagging tasks, was later adopted for chemical NER. ChemDataExtractor \cite{swain2016chemdataextractor} first introduced hybrid NLP pipelines combining rule-based and statistical components. 

With the emergence of transformer architectures, domain-specific pretrained models such as MatSciBERT \cite{gupta2022matscibert} and ChemBERTa \cite{chithrananda2020chemberta} demonstrated improved contextual understanding of scientific and chemical text. Building on these advances, Guo et al. \cite{guo2021automated} proposed a two-stage framework for reaction extraction from journal literature. In their approach, product extraction is formulated as a standard sequence labeling task using a transformer encoder followed by a CRF decoder under the BIO tagging scheme. For role labeling, reaction role entities are identified by inserting special markers around product spans to inform the encoder about the target product. For a given product, their model is able to detect eight associated reaction roles that are informative for describing chemical reactions. Recent work has explored data augmentation techniques to alleviate annotation scarcity in chemical reaction extraction. For example, Zhang et al. \cite{zhang2024chemical} and Dao et al. \cite{dao2025entity} employed synthetic sentence generation and LLM-based text augmentation. In contrast, our study examines whether active sample selection alone can improve annotation efficiency without introducing artificial data. 

\section{Methods}
\subsection{Model Architecture}
A typical active learning system iteratively improves a model by selecting informative unlabeled instances for annotation by an oracle and incorporating the newly labeled data into training\cite{settles2009active}.To fairly assess the effectiveness of applying active learning strategies to ChemBERT and ChemRxnBERT\cite{guo2021automated} models, we simulate pool-based active learning on the original annotated datasets\footnote{https://github.com/jiangfeng1124/ChemRxnExtractor/tree/main/tests/data} with identical data splits for both product extraction and role assignment. In both tasks, the data are organized as sentence-level blocks, where each block corresponds to a sentence extracted from a scientific article. Each sentence consists of a sequence of tokens annotated using the conventional BIO-tagging scheme. For product extraction, each token is assigned a single label indicating whether it belongs to a product entity(B-product or I-product) or not(O). For role labeling, each sentence may contain multiple label columns, each corresponding to the roles associated with a specific product mentioned in that sentence, resulting in 17 possible token labels(eight roles with B/I tags and O for not-a-role). 

Active learning is therefore performed at the sentence level. At the beginning of the process, all sentences from the training set are treated as unlabeled and form an unlabeled pool. In each round, a subset of sentences is selected according to an active learning strategy and sent for annotation. Once a sentence is picked, all of its token-level labels are revealed and added to the labeled training set. At every round, the labeled training set is reconstructed from the accumulated annotated sentence blocks, while the validation and test sets remain fixed throughout the entire process. Model training follows a warm-start strategy, with the first round initialized from the pretrained ChemBERT or ChemRxnBERT checkpoint, and subsequent rounds initialized from the checkpoint obtained in the previous round. 

\begin{figure*}[t]
  \centering
  \includegraphics[width=\textwidth]{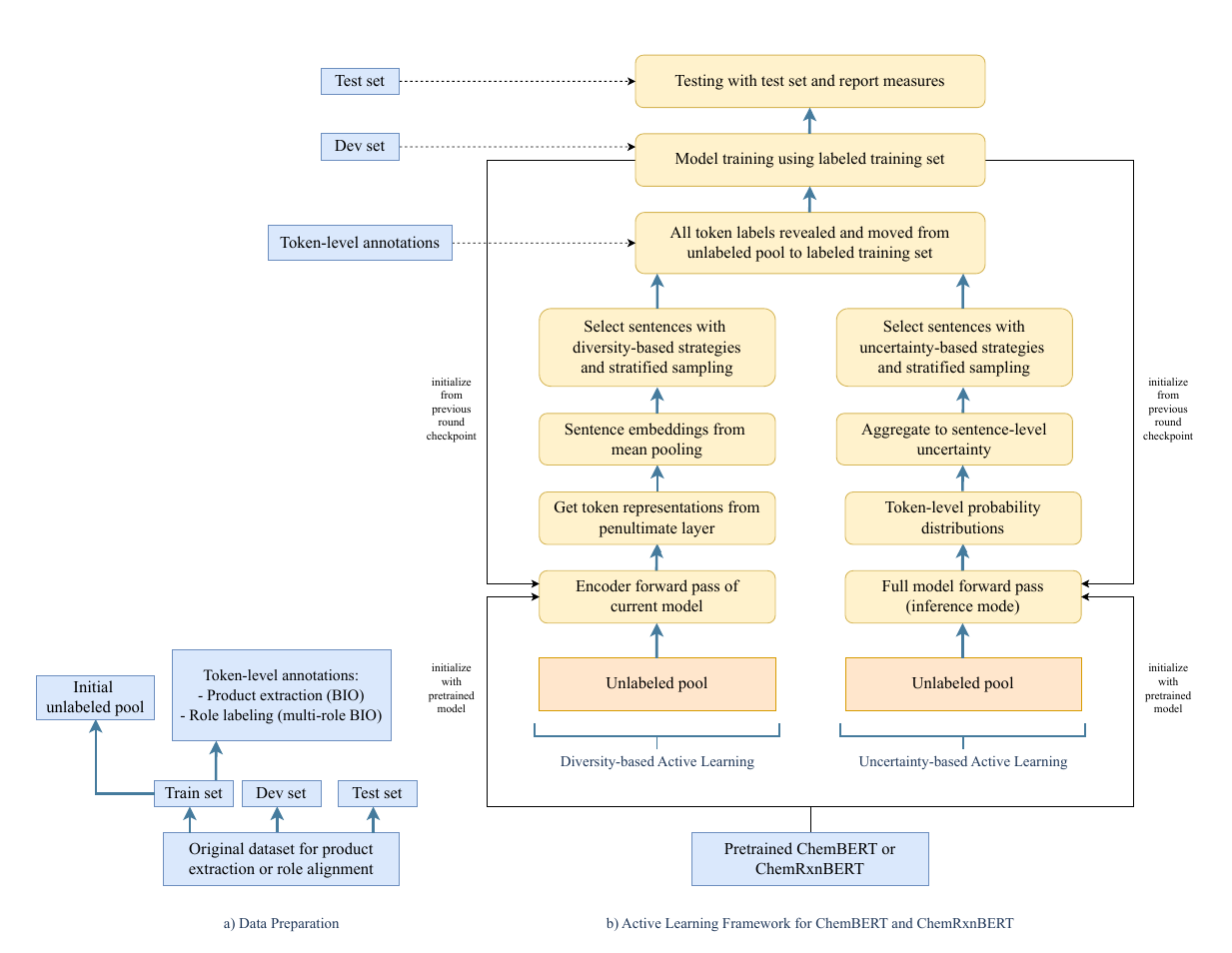}
  \caption{Pool-based active learning framework applied to product extraction based on ChemBERT and role labeling based on ChemRxnBERT. Although both tasks follow the same active learning procedure, they are trained and evaluated on different datasets.}
  \label{fig:al-framework}
\end{figure*}

We consider two classes of active learning strategies, which differ in their required inputs and querying mechanisms. Diversity-based methods rely on the properties of the input representations to select a diverse subset of unlabeled instances. Adapting to the ChemBERT and ChemRxnBert models, at each round, all unlabeled sentences are forwarded through the encoder of the current model, and token representations from the penultimate hidden layer are then aggregated into sentence embeddings by mean pooling. The resulting embeddings are used by diversity-based selection strategies to choose a batch of sentences for annotation.  In contrast, uncertainty-based methods operate on the prediction outputs of the model. The current model is applied in inference mode to the unlabeled pool to obtain token-level probability distributions, which are aggregated into sentence-level uncertainty scores based on the definition of uncertainty in each strategy. The scores are used to rank and select candidate sentences. In both cases, the selected sentences are fully annotated and added to the labeled training set, and the model is retrained before proceeding to the next round. The overall framework is illustrated in Figure~\ref{fig:al-framework}.

A key challenge in applying active learning to both tasks is the several label imbalance, with the majority of tokens assigned the O label. In product extraction, only a small fraction of sentences contain any product-related tokens, while in role assignment, most tokens are not associated with any role. In this scenario, simply applying active learning strategies can result in early rounds dominated by sentences containing exclusively O-labeled tokens, leading to degenerate models that only predict the majority class. To mitigate this issue, we incorporate stratified sampling into the selection process. For product extraction, unlabeled sentences are partitioned depending on whether they contain any product tokens, whereas for role assignment, sentences are grouped according to the number of distinct role types they contain. Active learning strategies are applied independently within each group, with selection quotas proportional to group sizes. This design preserves label distribution across rounds and ensures that informative non-O labels are not absent in the early stages of active learning, because of the preference of certain strategies. 

\subsection{Active Learning Strategies}
Seven active learning strategies are considered in this study, including two diversity-based approaches, four uncertainty-based techniques and random sampling as a baseline. 
Diversity based methods aim to select representative and diverse samples that can cover the unlabeled pool effectively. Rather than relying on model outputs, these methods operate on learned embeddings. 
\begin{itemize}
    \item The Core-set method\cite{sener2017active} is a classical diversity-driven approach in deep active learning, originally proposed for CNN. The core idea is to select a subset of unlabeled instances that best represent the overall data distribution by minimizing the distance between labeled and unlabeled samples. In contrast to the original implementation using the Euclidean distance, we compute cosine distance between embeddings of all samples, which is often more suitable for high-dimensional embeddings (in our case 768 dimension) and better captures semantic similarity. At each iteration, the unlabeled instance that maximizes the minimum cosine distance to the current labeled set is selected and added to the pool. The greedy selection continues until the desired number of samples is reached.
    \item CLUSTER+ is adapted from CLUSTER\cite{liu2024utilizing} with two modifications. First, the number of clusters is dynamically determined according to the size of the remaining unlabeled pool at each round, instead of using a fixed k in the original design. Let $U$ denote the number of unlabeled instances and $n$ the acquisition budget. We set the number of clusters to $k=\sqrt{U/2}$, constrained between 5 and $n$. The adaptive design prevents over-fragmentation when the pool becomes small. Second, we employ k-means++\cite{arthur2006k} instead of standard k-means to obtain more robust cluster initialization. The method first groups the sentence embeddings of the unlabeled pool with k-means++, then randomly draw approximately $n/k$ samples from each cluster.  
\end{itemize}

Uncertainty based techniques\cite{settles2009active} query instances for which the model is most uncertain about its predictions. Unlike diversity-driven approaches, they operate directly on the model's predictive distributions. As previous studies have shown, the effectiveness of different uncertainty methods depends on the specific application, and there is no universally optimal strategy\cite{wang2016cost, nachtegael2023study}. We therefore include the following approaches based on different uncertainty criteria. 
\begin{itemize}
    \item Least-confidence sampling\cite{settles2009active} selects instance whose most probable label is least confident. While it works well in fields like image classification\cite{wang2016cost}, standard least-confidence sampling has been shown to prefer longer sentences in sequence labeling tasks\cite{liu2020ltp, agrawal2021active}. To mitigate this bias, we adopt a modified formulation inspired by\cite{agrawal2021active}, that takes into account the sentence lengths. The modified formulation is primarily beneficial in settings with larger label spaces. For product extraction, which involves only three BIO tags, we use the standard least-confidence definition. In role labeling task, 
    for a sentence s, the uncertainty is 
    \begin{equation}
        s_{MLC} = \frac{\sum_{i=1}^N 1-\ P_\theta(\hat{y_1}|x_i)}{N},
    \end{equation}
    where $\hat{y_1}$ refers to the most probable label of the i-th least confident token $x_i$. Instead of using several fixed N in the original experiments\cite{agrawal2021active}, we set $N=\sqrt{L/2}$, where $L$ denotes the sentence lengths. The sentences with the highest scores are selected.  
    \item Margin sampling\cite{scheffer2001active} measures the ambiguity between the two most likely labels. The margin for a token $x $ is defined as
    \begin{equation}
        x_M = P_{\theta}(\hat{y_1}|x) - P_{\theta}(\hat{y_2}|x),
    \end{equation}
    where $\hat{y_1}$ and $\hat{y_2}$ denote the first and second most probable labels. A smaller difference indicates higher uncertainty. After aggregation, sentences with the smallest margins are selected.  
    \item Entropy sampling\cite{settles2009active} quantifies uncertainty by considering the full predictive distribution. For token $x$, entropy is computed as
    \begin{equation}
        x_H =  - \sum_i P_{\theta}(y_i|x) log P_{\theta}(y_i|x),
    \end{equation}
    where $y_i$ represents all possible labels. Higher entropy indicates greater uncertainty in labels. In practice, tiny probabilities are removed to avoid numerical instability. Sentence-level scores are derived by aggregating token-level entropy. 

\item BALD-based batch selection employs Bayesian active learning by disagreements(BALD)\cite{houlsby2011bayesian} to estimate uncertainty independently for each sentence, and perform batch construction externally via stratified selection. BALD selects instances for which the model parameters under the posterior disagree the most about the outcome. Following previous works\cite{gal2016dropout,gal2017deep}, we approximate Bayesian inference using Monte Carlo (MC) dropout, performing multiple stochastic forward passes with dropout enabled. The uncertainty of a token is quantified as the mutual information between predictions and model parameters, more specifically, as the difference between the entropy of the mean prediction and the expected entropy across MC samples. For product extraction, token-level scores are aggregated to the sentence level. For role extraction, uncertainty is computed over emission probabilities conditioned on given product spans and aggregated at the sequence level. Although BatchBALD\cite{kirsch2019batchbald} extends BALD to joint batch acquisition by maximizing mutual information over subsets of instances, it is not applicable in our setting, due to the stratified sampling constraints to preserve label distributions. 
\end{itemize}
Finally, random selection serves as a baseline, simulating passive learning by selecting instances uniformly at random. For fair comparison, stratified sampling is applied in the same manner as for the other strategies to maintain consistent label distributions across rounds.

\section{Results}
\subsection{Implementation and evaluation}
 Following the original training setup, we use ChemBERT for product extraction with batch size 16, maximum input sequence length 256 and CRF learning rate 5e-3, trained for 2 epochs per active learning round. For role labeling, we employ ChemRxnBERT with batch size 6 and maximum sequence length 512, while keeping the same epoch times and learning rate. The product extraction dataset contains 6163/698/723 sentences for training/validation/testing, while role labeling dataset has 387/57/67 sentence blocks. Considering the size of training set, we simulate pool-based active learning over 10 rounds with a fixed acquisition budget of 10\% of the original training pool per round (rounded to at least one sentence). 
 For uncertainty-based methods, token-level predictive distributions are computed in inference mode and converted to probabilities, including exponentiating log-normalized emissions for product extraction and applying softmax to logits for role labeling. Token uncertainties are aggregated into sentence-level scores by averaging over valid decoding positions, excluding padding and special tokens. 

 Accuracy, precision and F1 score are selected as they are dominantly used for evaluating the performance of active learning strategies\cite{kath2026the}. Using budget as the stopping criterion, we compare the performance metrics with the random baseline and the reported passive learning results\cite{guo2021automated} for each round. 

\subsection{Results discussion}
\begin{table*}[t]
\centering
\caption{Test F1 (\%) across rounds for product extraction. The best value in each strategy is in bold. The passive learning baseline is 76.24. MLC stands for modified least confidence sampling. BALD-batch represents BALD-based batch selection. The representations remain in the following graphs and tables. }
\label{tab:prod_f1_rounds}
\begin{adjustbox}{max width=\textwidth}
\begin{tabular}{lcccccccccc}
\toprule
Strategy $\backslash$ Round & 1 & 2 & 3 & 4 & 5 & 6 & 7 & 8 & 9 & 10 \\
\midrule
Core-set & 0.00 & 6.84 & 58.24 & 64.21 & 67.63 & 72.36 & \textbf{75.44} & 71.03 & 70.09 & 65.98 \\
CLUSTER+ & 38.67 & 70.59 & 73.91 & 72.65 & 69.83 & \textbf{74.65} & 73.78 & 71.49 & 69.65 & 69.35 \\
BALD-batch & 67.31 & 56.98 & 61.71 & 56.32 & 60.57 & 64.48 & 63.33 & 60.11 & 73.27 & \textbf{73.58} \\
MLC & 60.51 & 57.92 & 55.43 & 55.68 & 64.52 & 63.33 & 63.39 & 73.02 & \textbf{74.75} & 68.90 \\
Margin & 58.17 & 58.06 & 62.65 & 63.64 & 62.43 & 70.21 & 69.43 & 68.02 & \textbf{72.12} & 70.53 \\
Entropy & 63.96 & 54.65 & 50.60 & 49.37 & 62.89 & 67.01 & 60.00 & 66.67 & \textbf{69.61} & 68.84 \\
Random & 67.00 & 68.90 & \textbf{72.10} & 66.97 & 68.81 & 70.09 & 68.60 & 71.50 & 67.35 & 68.00 \\
\bottomrule
\end{tabular}
\end{adjustbox}
\end{table*}

\begin{table*}[t]
\centering
\caption{Test F1 (\%) across rounds for role labeling. The best value in each strategy is in bold. The passive learning baseline is 78.66. }
\label{tab:role_f1_rounds}
\begin{adjustbox}{max width=\textwidth}
\begin{tabular}{lcccccccccc}
\toprule
Strategy $\backslash$ Round & 1 & 2 & 3 & 4 & 5 & 6 & 7 & 8 & 9 & 10 \\
\midrule
Core-set & 0.00 & 54.44 & 66.67 & 70.91 & 73.44 & 76.15 & \textbf{78.60} & 76.10 & 76.38 & 76.83 \\
CLUSTER+ & 4.08 & 57.19 & 65.77 & 67.44 & 74.66 & 76.83 & 76.18 & 76.71 & 76.22 & \textbf{78.54} \\
BALD-batch & 32.82 & 59.58 & 67.45 & 71.62 & 71.18 & 74.26 & 74.74 & 74.48 & \textbf{76.02} & 75.11 \\
MLC & 0.00 & 51.44 & 63.95 & 71.43 & 73.65 & 75.34 & 77.16 & 77.41 & \textbf{77.89} & 76.58 \\
Margin & 2.90 & 52.68 & 64.16 & 67.85 & 74.10 & 74.81 & 75.57 & 74.81 & \textbf{76.90} & 76.15 \\
Entropy & 0.00 & 44.24 & 60.00 & 69.16 & 75.56 & \textbf{77.18} & 75.72 & 75.95 & 77.09 & 76.69 \\
Random & 0.00 & 30.00 & 54.63 & 65.68 & 69.33 & 68.99 & 73.09 & 73.58 & \textbf{75.81} & 74.12 \\
\bottomrule
\end{tabular}
\end{adjustbox}
\end{table*}

\begin{figure*}[p]
\centering
\begin{subfigure}{0.48\textwidth}
  \centering
  \includegraphics[width=\linewidth]{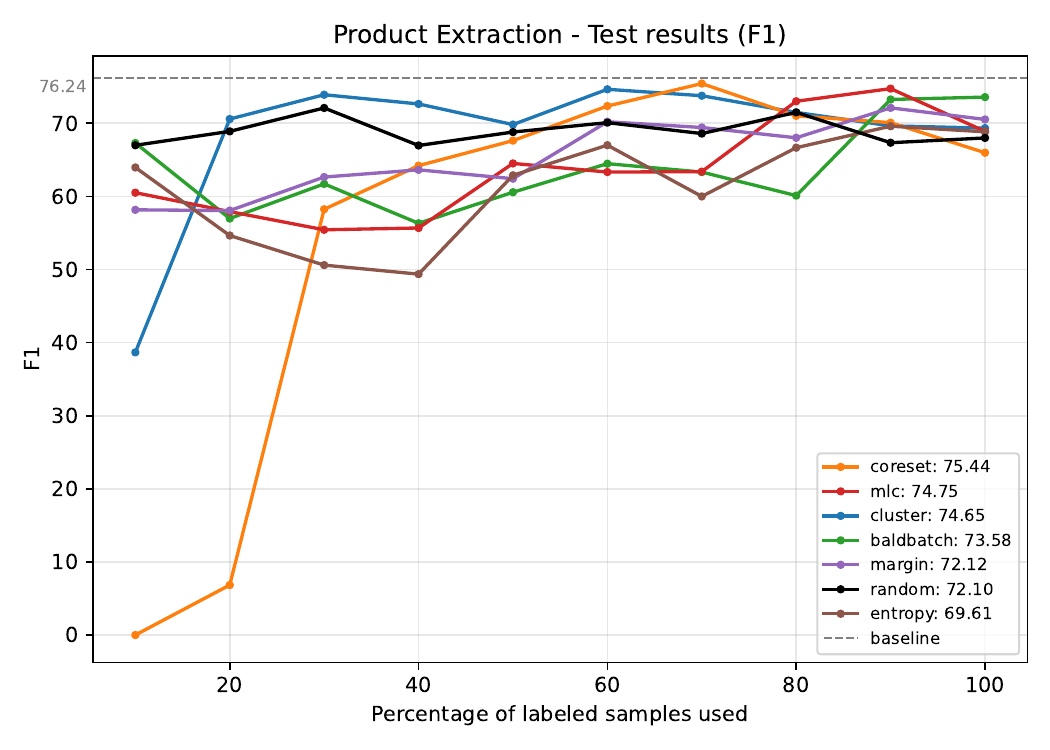}
  \caption{Product extraction F1}
\end{subfigure}\hfill%
\begin{subfigure}{0.48\textwidth}
  \centering
  \includegraphics[width=\linewidth]{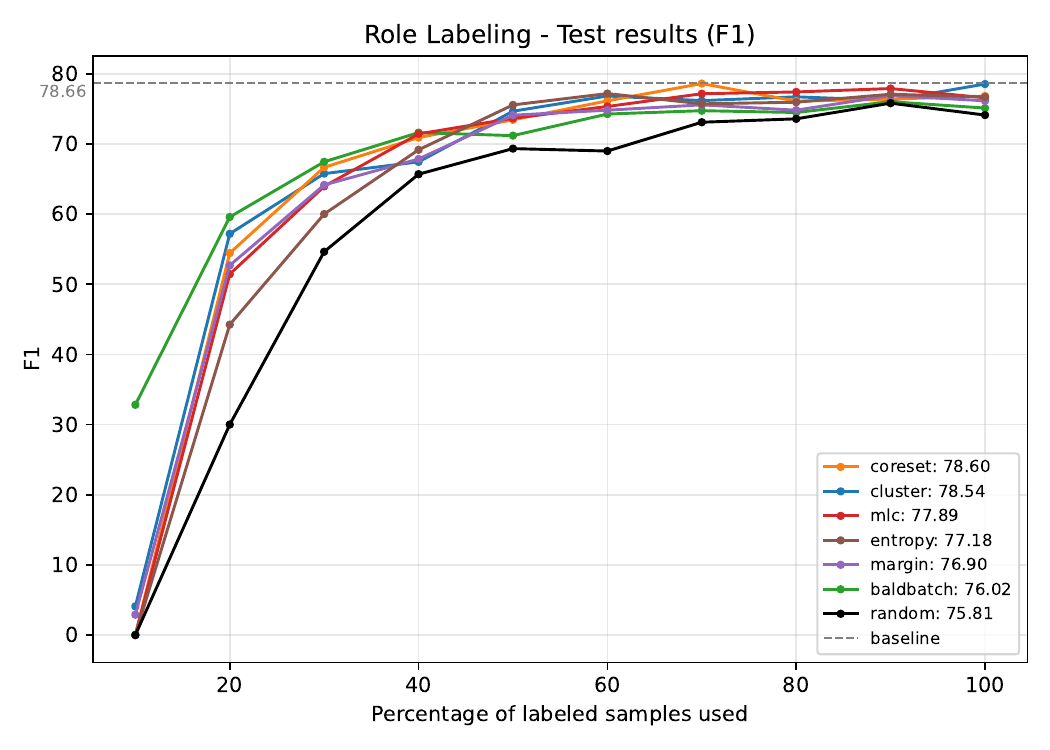}
  \caption{Role labeling F1}
\end{subfigure}\hfill%

\vspace{4pt}

\begin{subfigure}{0.48\textwidth}
  \centering
  \includegraphics[width=\linewidth]{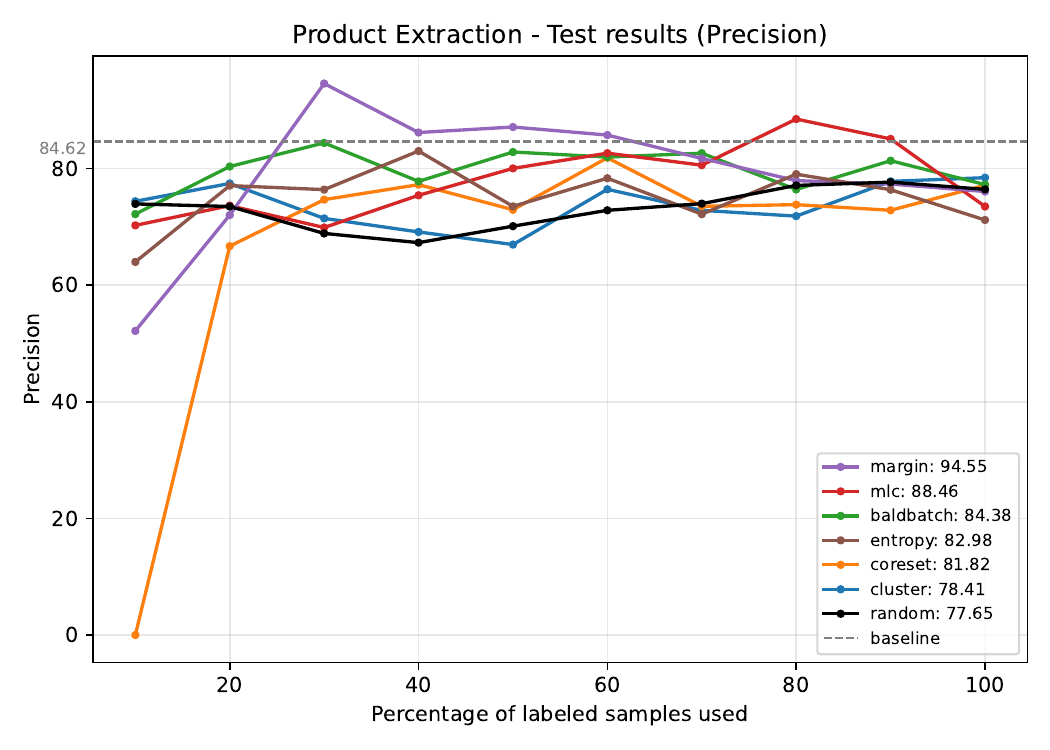}
  \caption{Product extraction precision}
\end{subfigure}\hfill%
\begin{subfigure}{0.48\textwidth}
  \centering
  \includegraphics[width=\linewidth]{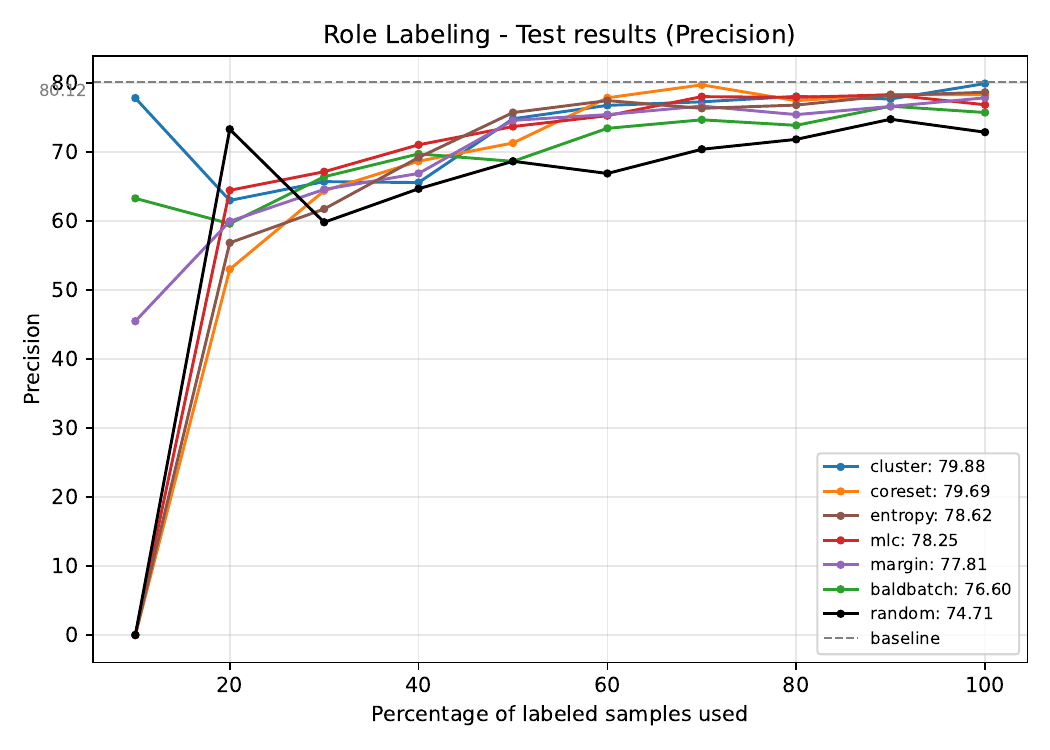}
  \caption{Role labeling precision}
\end{subfigure}

\vspace{4pt}

\begin{subfigure}{0.48\textwidth}
  \centering
  \includegraphics[width=\linewidth]{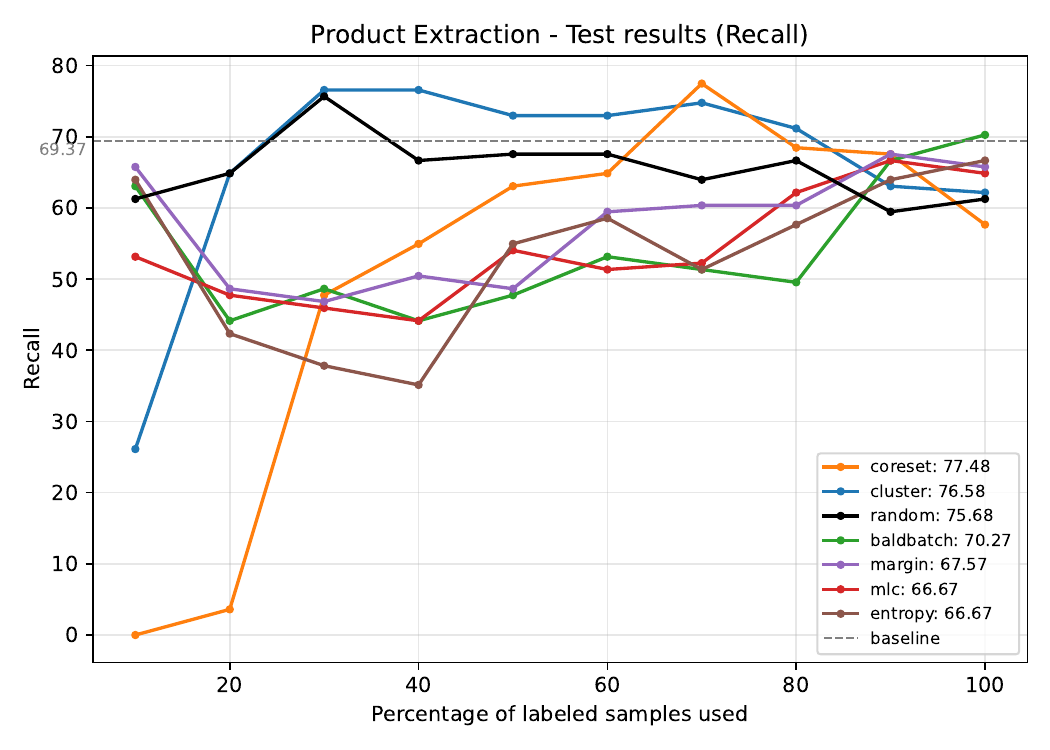}
  \caption{Product extraction recall}
\end{subfigure}\hfill%
\begin{subfigure}{0.48\textwidth}
  \centering
  \includegraphics[width=\linewidth]{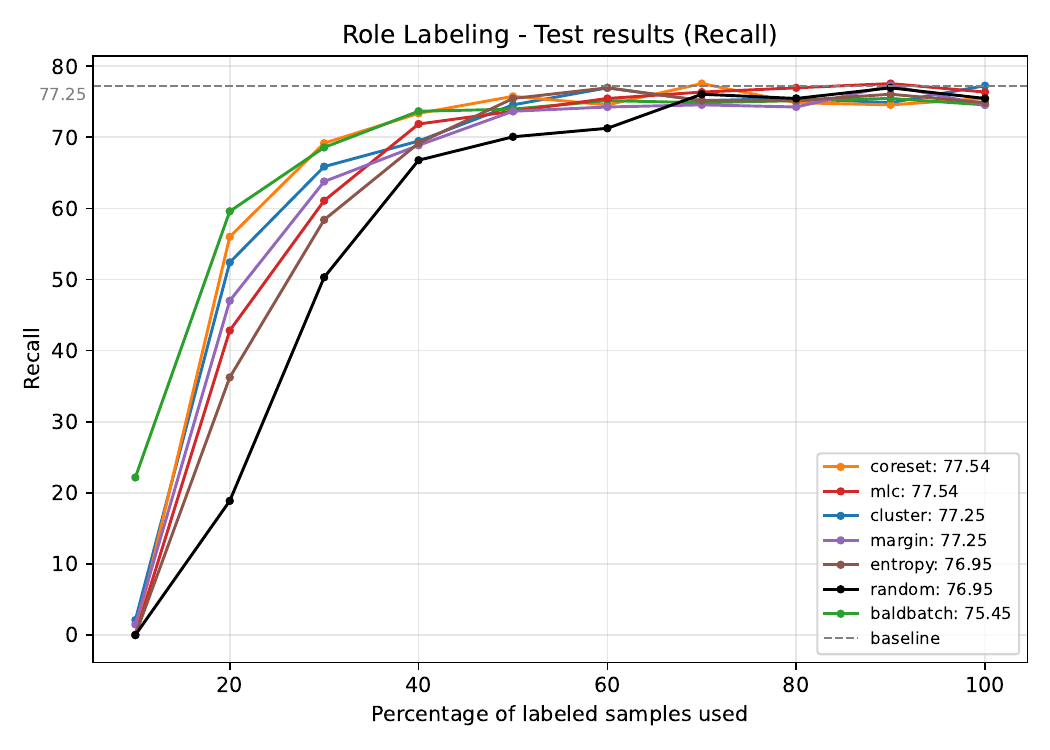}
  \caption{Role labeling recall}
\end{subfigure}
\caption{The learning curves of each active learning strategy. The dash line refers to the reported passive learning results\cite{guo2021automated}. The legend is sorted by the peak value of each strategy.}
\label{fig:learning_curve}
\end{figure*}

As shown in Figure~\ref{fig:learning_curve}, the learning curves illustrate active learning performance across metric against the fraction of labeled training data. For product extraction, performance varies considerably across strategies, particularly in the early rounds. The Core-set method achieves the highest F1 score among all strategies when using 70\% of the annotated data, while the best performance of CLUSTER+ is reached at round 6. In contrast, uncertainty-based methods display more fluctuating trajectories, lacking consistency across rounds. Despite incorporating stratified sampling to preserve positive instances across rounds, performance remains unstable and shows a pronounced precision–recall trade-off. Strategies often attain high precision at the expense of low recall, implying that later acquisitions may lead the model to make increasingly conservative predictions with the risk of missing valid product entities. 
The unstable performance is likely related to the sparsity of product mentions. Even with stratified sampling at the sentence level, token-level label imbalance persists and changes in the structure of labeled pool could be influential. This may also explain why random sampling performs competitively with other active acquisition until the final rounds.

In role labeling, the learning curves are smoother and more tightly clustered across strategies, with most methods experiencing relatively steady improvement throughout training. The Core-set method reaches near-baseline performance using only 70\% of the full labeled dataset, followed by entropy sampling showing a performance difference of less than 1.5\% from the baseline as early as round 6. While greater differences appear in the first half of the active learning process,  the gaps between active learning strategies narrow as additional labeled data are acquired. It suggests that for this task model performance benefits more from the simple accumulation of supervision, than from the specific selection strategy. Nevertheless, all active learning selection outperform random sampling consistently. Precision and recall tend to increase simultaneously and stabilize in later rounds, in contrast to the marked tradeoff observed in product extraction. Such behavior aligns with the nature of role labeling task, where a larger tag set and sequence constraints promote gradual improvement and reduce sensitivity to the exact composition of the accumulated labeled pool. The F1 values across rounds for both tasks are presented in Table~\ref{tab:prod_f1_rounds} and Table~\ref{tab:role_f1_rounds}. The corresponding precision and recall results are provided in the Appendix (Tables~\ref{tab:prod_precision_rounds}, \ref{tab:prod_recall_rounds}, \ref{tab:role_precision_rounds} and \ref{tab:role_recall_rounds}) due to space limit. 

Overall, although none of the active learning strategies consistently surpass full-data passive training, they are able to approach comparable performance using fewer labeled instances. The non-monotonic learning curves observed in both tasks can be partly attributed to distribution shift introduced during iterative acquisition and warm-start training. In addition, severe token-level class imbalance may amplify sampling bias across rounds, particularly for product extraction. While early rounds tend to capture either representative or uncertain samples, later-round query may increasingly contain ambiguous or noisy sentences that contribute limited benefit and even lead to poorer generalization. Moreover, when strong contextual representations are already learned during pretraining of the encoders, selecting informative instances may not necessarily produce performance gains. Meanwhile, the CRF layer performs decoding with transition constraints, but uncertainty computed at sentence level may not fully capture the enforced token-level constraints. 

\subsection{Selection preference}

The two selected diversity-driven strategies reach higher peak performance than uncertainty-based methods in both tasks. However, especially for product extraction, their performance declines significantly in later rounds. To understand this and further analyze the behavior of diversity-based methods, we visualize their selection patterns in embedding space. As described earlier, sentence embeddings are extracted from the encoder at each round as the input for active selection, which are 768-dimensional vectors. We project them into two dimensions using t-Distributed Stochastic Neighbor Embedding (t-SNE) \cite{van2008visualizing}. Selected samples are highlighted to illustrate how each strategy explores the feature space, and for CLUSTER+, detected clusters are also shown. The larger size of the product extraction dataset enables clearer display of the pattern compared with the role labeling task. 
\clearpage
\clearpage

\begin{figure*}[t]
\centering
\begin{subfigure}{0.32\textwidth}
  \centering
  \includegraphics[width=\linewidth]{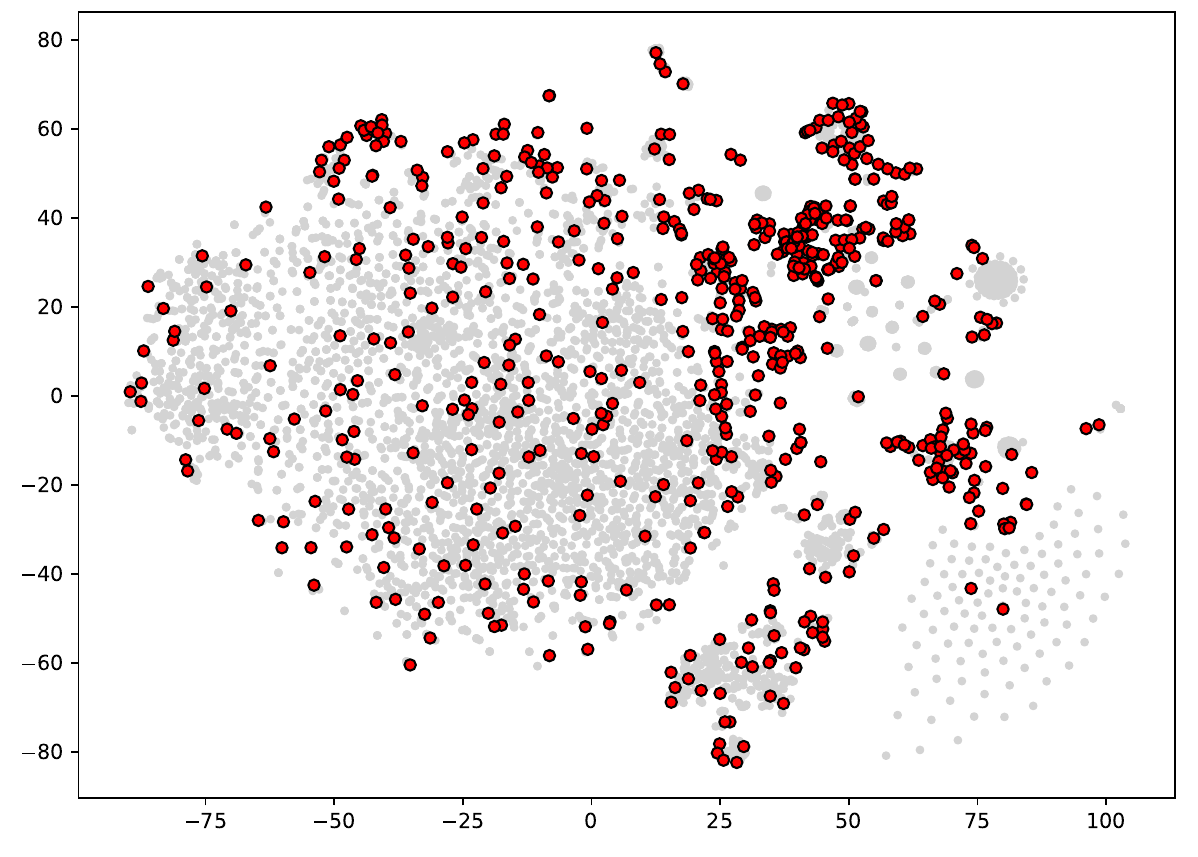}
  \caption{Core-set round 1}
\end{subfigure}\hfill%
\begin{subfigure}{0.32\textwidth}
  \centering
  \includegraphics[width=\linewidth]{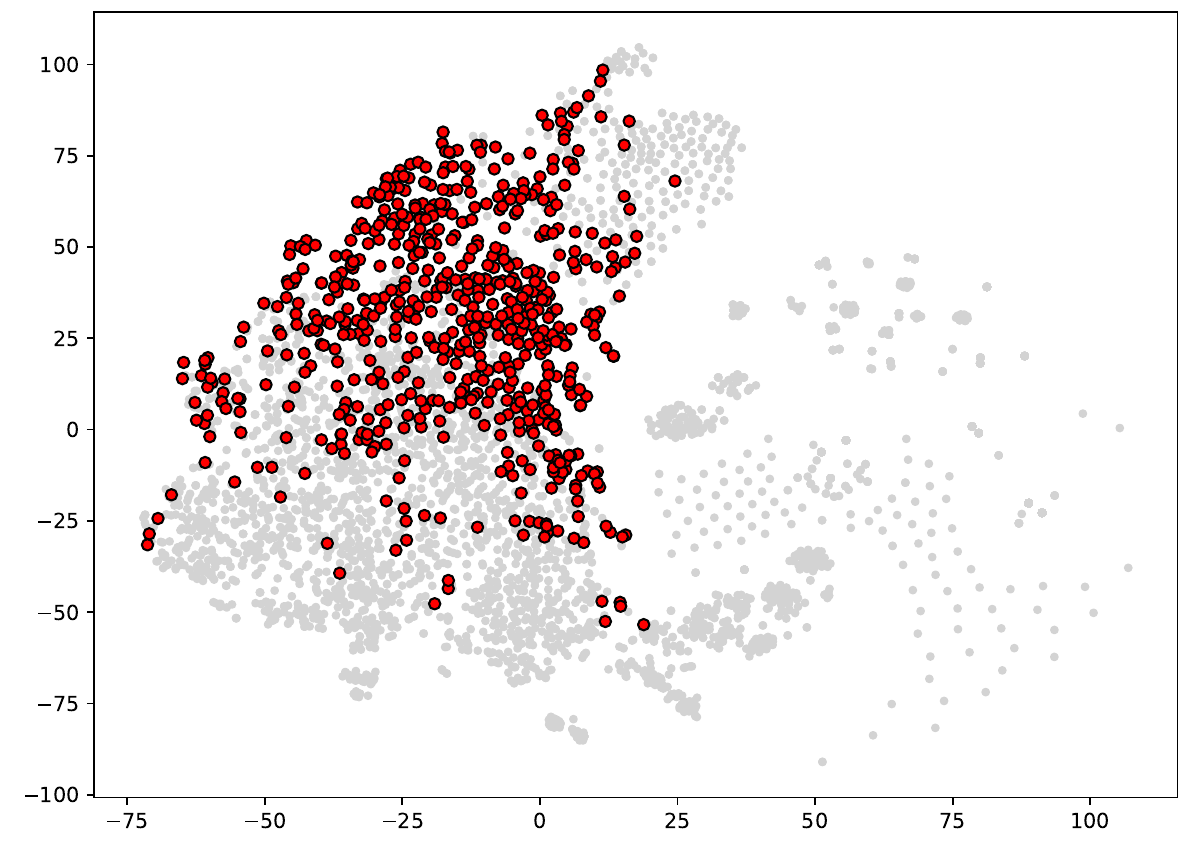}
  \caption{Core-set round 4}
\end{subfigure}\hfill%
\begin{subfigure}{0.32\textwidth}
  \centering
  \includegraphics[width=\linewidth]{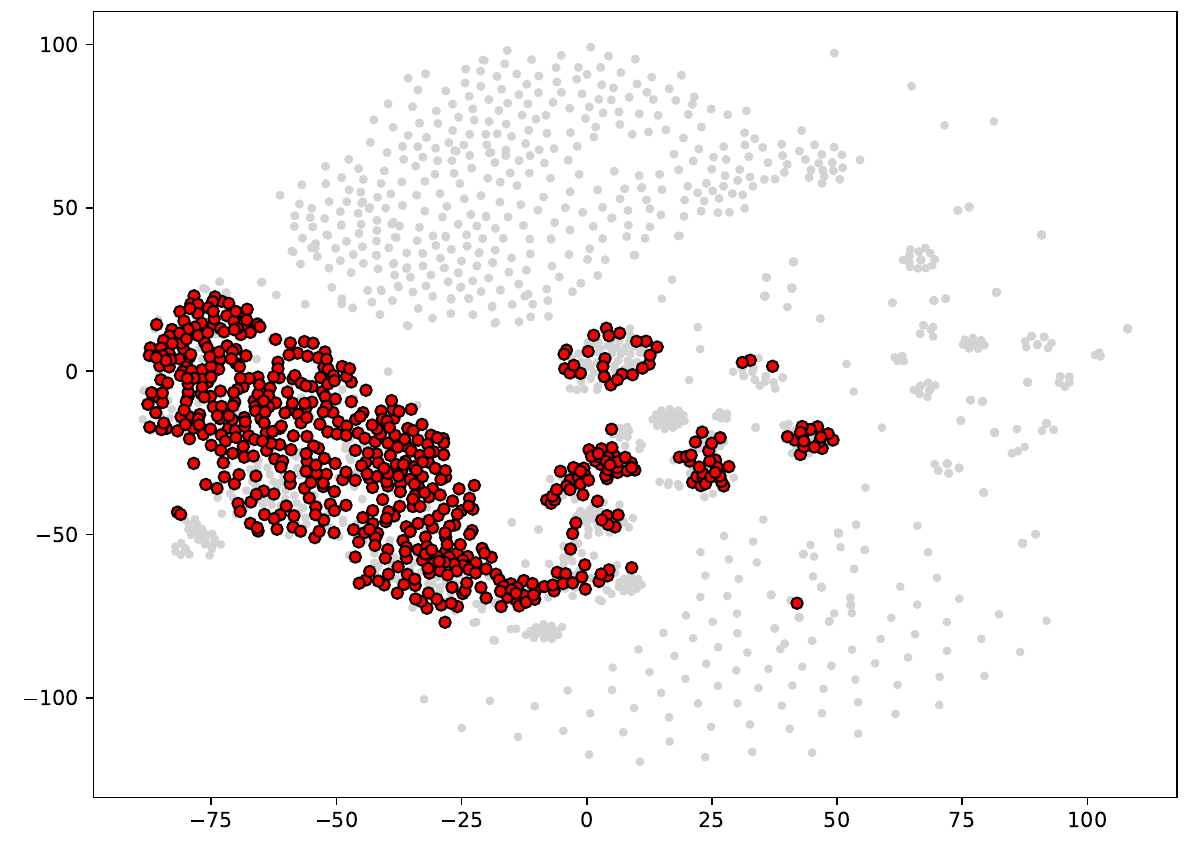}
  \caption{Core-set round 7}
\end{subfigure}

\vspace{4pt}

\begin{subfigure}{0.32\textwidth}
  \centering
  \includegraphics[width=\linewidth]{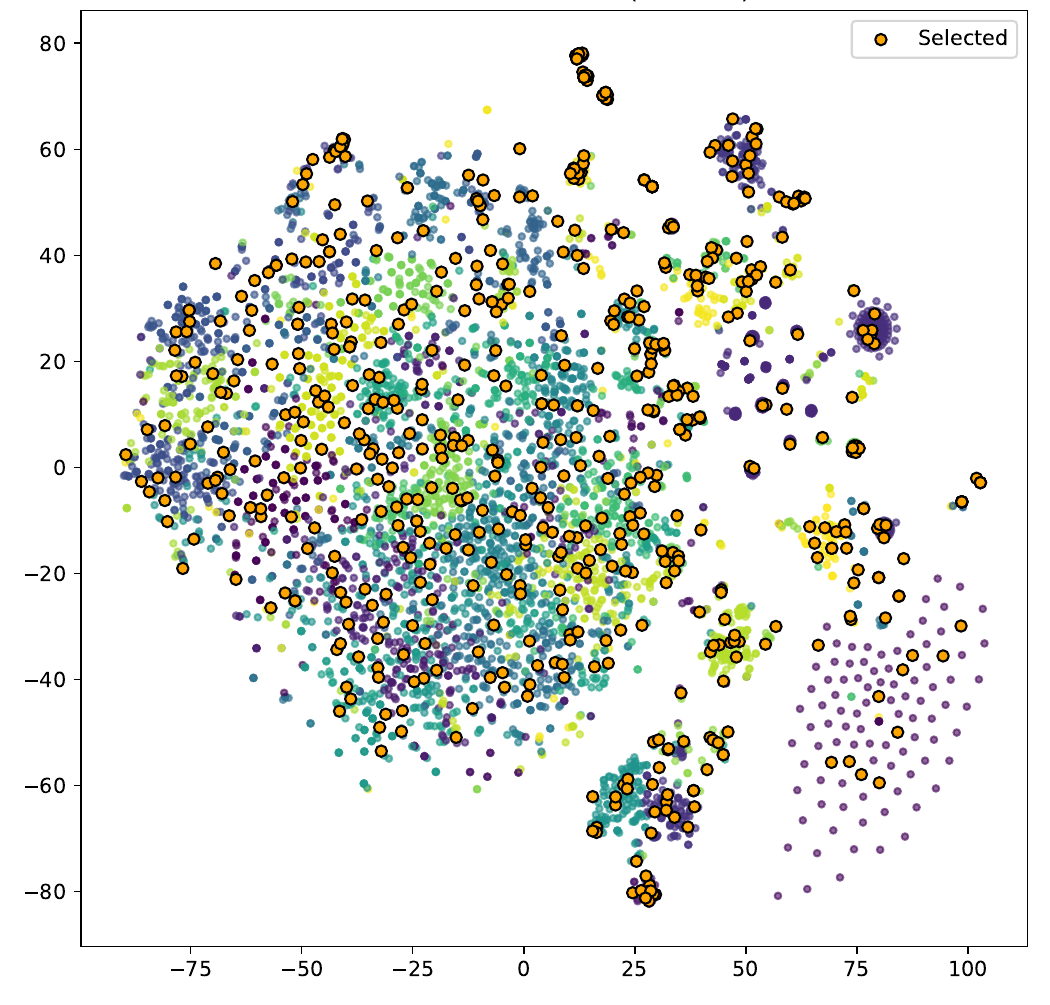}
  \caption{CLUSTER+ round 1}
\end{subfigure}\hfill%
\begin{subfigure}{0.32\textwidth}
  \centering
  \includegraphics[width=\linewidth]{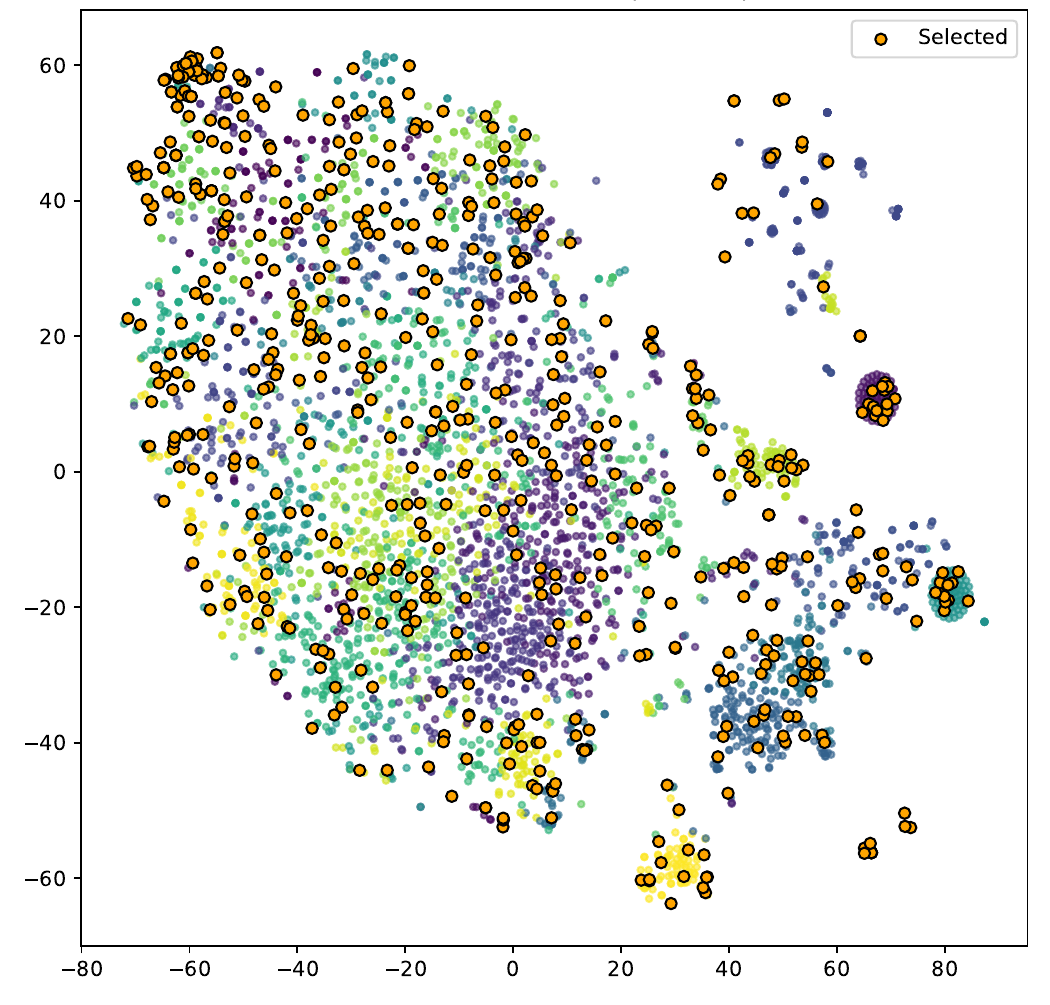}
  \caption{CLUSTER+ round 4}
\end{subfigure}\hfill%
\begin{subfigure}{0.32\textwidth}
  \centering
  \includegraphics[width=\linewidth]{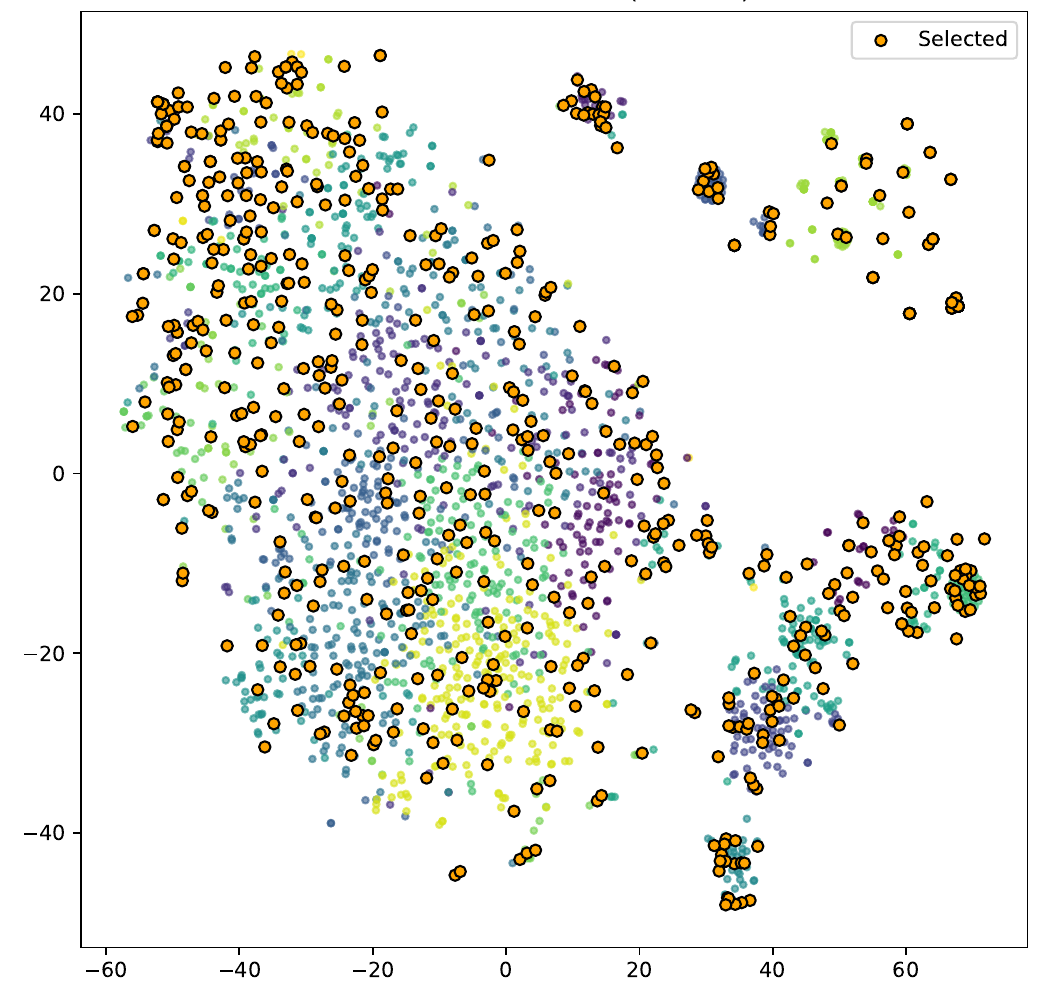}
  \caption{CLUSTER+ round 7}
\end{subfigure}
\caption{Visualization of sample selection process of Core-set and CLUSTER+ strategies for product extraction across rounds.}
\label{fig:selection_process}
\end{figure*}
Figure~\ref{fig:selection_process} presents the selection process of Core-set and CLUSTER+ at round 1, 4 and 7. Samples selected by Core-set appear relatively concentrated within specific regions of the embedding space. This suggests that Core-set may prioritize refining already sampled areas rather than broadly exploring the space. In contrast, CLUSTER+ selects instances from multiple regions, including smaller and less populated clusters, which are largely overlooked by Core-set. These differences become more pronounced as active learning progresses. By round 7, the remaining unlabeled pool exhibits great difference in spatial distribution depending on the strategy applied. Core-set tends to leave peripheral and sparse regions under-explored, while CLUSTER+ maintains a more balanced coverage of the feature space. The observation indicates that diversity-driven active learning strategies affect not only the order of selecting samples, but also the structure of the remaining unlabeled pool, which in turn may influence the performance of the model across rounds. 
Once early rounds shape the labeled pool toward certain regions of the embedding space, subsequent warm-start training can amplify these biases, possibly leading to mid-round peaks followed by performance decline. 

\section{Conclusion and Future Work}
In this study, we investigate the integration of uncertainty-based and diversity-based active learning strategies with ChemBERT and ChemRxnBERT for product extraction and reaction role labeling. While none of the active learning strategies consistently surpass passive learning on the full dataset, certain strategies such as Core-set achieve competitive performance with reduced labeled data. However, the learning curves are often non-monotonic, typically reaching their peak performance in the middle rounds and declining thereafter. Our analysis suggests that the influence of active learning is closely tied to model architecture and task characteristics. For pretrained transformer encoders coupled with CRF decoders, strong contextual representations and structured decoding constraints may reduce the effectiveness of uncertainty-based active acquisition. Additionally, persistent token-level imbalance and warm-start training can amplify sampling bias across rounds. For diversity-based strategies, this means that the structure of the remaining unlabeled pool may be severely affected and become biased. These factors limit the stability and consistency of active learning gains.

Future work may explore active learning strategies tailored to the model architecture and task features. Strategies explicitly accounting for distribution shifts across rounds may also be promising. To mitigate over-representation of noisy data in later rounds, acquisition could be made more balanced from the beginning, such as combining multiple criteria during one round of selection, and selecting the most uncertain samples within clusters. Furthermore, the unlabeled pool could be progressively filtered by taking noise into account, rejecting samples that are likely noisy or uninformative in later rounds.

\clearpage
\bibliography{references}

@article{sang2003introduction,
  title={Introduction to the CoNLL-2003 shared task: Language-independent named entity recognition},
  author={Sang, Erik F and De Meulder, Fien},
  journal={arXiv preprint cs/0306050},
  year={2003}
}

@article{jehangir2023survey,
  title={A survey on named entity recognition—datasets, tools, and methodologies},
  author={Jehangir, Basra and Radhakrishnan, Saravanan and Agarwal, Rahul},
  journal={Natural Language Processing Journal},
  volume={3},
  pages={100017},
  year={2023},
  publisher={Elsevier}
}

@article{weischedel2011ontonotes,
  title={Ontonotes release 4.0},
  author={Weischedel, Ralph and Pradhan, Sameer and Ramshaw, Lance and Palmer, Martha and Xue, Nianwen and Marcus, Mitchell and Taylor, Ann and Greenberg, Craig and Hovy, Eduard and Belvin, Robert and others},
  journal={LDC2011T03, Philadelphia, Penn.: Linguistic Data Consortium},
  volume={17},
  year={2011}
}

@article{shen2017deep,
  title={Deep active learning for named entity recognition},
  author={Shen, Yanyao and Yun, Hyokun and Lipton, Zachary C and Kronrod, Yakov and Anandkumar, Animashree},
  journal={arXiv preprint arXiv:1707.05928},
  year={2017}
}

@article{ren2021survey,
  title={A survey of deep active learning},
  author={Ren, Pengzhen and Xiao, Yun and Chang, Xiaojun and Huang, Po-Yao and Li, Zhihui and Gupta, Brij B and Chen, Xiaojiang and Wang, Xin},
  journal={ACM computing surveys (CSUR)},
  volume={54},
  number={9},
  pages={1--40},
  year={2021},
  publisher={ACM New York, NY}
}

@inproceedings{yang2016active,
  title={Active learning using uncertainty information},
  author={Yang, Yazhou and Loog, Marco},
  booktitle={2016 23rd International Conference on Pattern Recognition (ICPR)},
  pages={2646--2651},
  year={2016},
  organization={IEEE}
}

@article{wang2016cost,
  title={Cost-effective active learning for deep image classification},
  author={Wang, Keze and Zhang, Dongyu and Li, Ya and Zhang, Ruimao and Lin, Liang},
  journal={IEEE Transactions on Circuits and Systems for Video Technology},
  volume={27},
  number={12},
  pages={2591--2600},
  year={2016},
  publisher={IEEE}
}

@inproceedings{gal2017deep,
  title={Deep bayesian active learning with image data},
  author={Gal, Yarin and Islam, Riashat and Ghahramani, Zoubin},
  booktitle={International conference on machine learning},
  pages={1183--1192},
  year={2017},
  organization={PMLR}
}

@article{shelmanov2021active,
  title={Active learning for sequence tagging with deep pre-trained models and Bayesian uncertainty estimates},
  author={Shelmanov, Artem and Puzyrev, Dmitri and Kupriyanova, Lyubov and Belyakov, Denis and Larionov, Daniil and Khromov, Nikita and Kozlova, Olga and Artemova, Ekaterina and Dylov, Dmitry V and Panchenko, Alexander},
  journal={arXiv preprint arXiv:2101.08133},
  year={2021}
}

@article{sener2017active,
  title={Active learning for convolutional neural networks: A core-set approach},
  author={Sener, Ozan and Savarese, Silvio},
  journal={arXiv preprint arXiv:1708.00489},
  year={2017}
}

@article{liu2024utilizing,
  title={Utilizing active learning strategies in machine-assisted annotation for clinical named entity recognition: a comprehensive analysis considering annotation costs and target effectiveness},
  author={Liu, Jiaxing and Wong, Zoie SY},
  journal={Journal of the American Medical Informatics Association},
  volume={31},
  number={11},
  pages={2632--2640},
  year={2024},
  publisher={Oxford University Press}
}

@article{li2022ud_bbc,
  title={UD\_BBC: Named entity recognition in social network combined BERT-BiLSTM-CRF with active learning},
  author={Li, Wei and Du, Yajun and Li, Xianyong and Chen, Xiaoliang and Xie, Chunzhi and Li, Hui and Li, Xiaolei},
  journal={Engineering Applications of Artificial Intelligence},
  volume={116},
  pages={105460},
  year={2022},
  publisher={Elsevier}
}

@inproceedings{liu2024cybersecurity,
  title={A cybersecurity named entity recognition model based on active learning and self-learning},
  author={Liu, Zhaoli and Jiang, Kun and Liu, Zheng and Qin, Tao},
  booktitle={2024 36th Chinese Control and Decision Conference (CCDC)},
  pages={4505--4510},
  year={2024},
  organization={IEEE}
}

@article{guo2021automated,
  title={Automated chemical reaction extraction from scientific literature},
  author={Guo, Jiang and Ibanez-Lopez, A Santiago and Gao, Hanyu and Quach, Victor and Coley, Connor W and Jensen, Klavs F and Barzilay, Regina},
  journal={Journal of chemical information and modeling},
  volume={62},
  number={9},
  pages={2035--2045},
  year={2021},
  publisher={ACS Publications}
}

@article{kirsch2019batchbald,
  title={Batchbald: Efficient and diverse batch acquisition for deep bayesian active learning},
  author={Kirsch, Andreas and Van Amersfoort, Joost and Gal, Yarin},
  journal={Advances in neural information processing systems},
  volume={32},
  year={2019}
}

@article{settles2009active,
  title={Active learning literature survey},
  author={Settles, Burr},
  year={2009},
  publisher={University of Wisconsin-Madison Department of Computer Sciences}
}

@misc{reaxys,
  author = {},
  title = {},
  year = {},
  note ={Reaxys – An expert-curated chemistry database. \url{https://www.elsevier.com/products/reaxys}}}

@misc{scifinder,
  author = {},
  title = {},
  year = {},
  note ={CAS SciFinder - Chemical Compound Database. \url{https://www.cas.org/solutions/cas-scifinder-discovery-platform}}}

@article{sadybekov2023computational,
  title={Computational approaches streamlining drug discovery},
  author={Sadybekov, Anastasiia V and Katritch, Vsevolod},
  journal={Nature},
  volume={616},
  number={7958},
  pages={673--685},
  year={2023},
  publisher={Nature Publishing Group UK London}
}

@article{hawizy2011chemicaltagger,
  title={ChemicalTagger: A tool for semantic text-mining in chemistry},
  author={Hawizy, Lezan and Jessop, David M and Adams, Nico and Murray-Rust, Peter},
  journal={Journal of cheminformatics},
  volume={3},
  number={1},
  pages={17},
  year={2011},
  publisher={Springer}
}

@article{coley2019graph,
  title={A graph-convolutional neural network model for the prediction of chemical reactivity},
  author={Coley, Connor W and Jin, Wengong and Rogers, Luke and Jamison, Timothy F and Jaakkola, Tommi S and Green, William H and Barzilay, Regina and Jensen, Klavs F},
  journal={Chemical science},
  volume={10},
  number={2},
  pages={370--377},
  year={2019},
  publisher={Royal Society of Chemistry}
}

@article{gu2021domain,
  title={Domain-specific language model pretraining for biomedical natural language processing},
  author={Gu, Yu and Tinn, Robert and Cheng, Hao and Lucas, Michael and Usuyama, Naoto and Liu, Xiaodong and Naumann, Tristan and Gao, Jianfeng and Poon, Hoifung},
  journal={ACM Transactions on Computing for Healthcare (HEALTH)},
  volume={3},
  number={1},
  pages={1--23},
  year={2021},
  publisher={ACM New York, NY}
}

@article{zhang2024fine,
  title={Fine-tuning large language models for chemical text mining},
  author={Zhang, Wei and Wang, Qinggong and Kong, Xiangtai and Xiong, Jiacheng and Ni, Shengkun and Cao, Duanhua and Niu, Buying and Chen, Mingan and Li, Yameng and Zhang, Runze and others},
  journal={Chemical science},
  volume={15},
  number={27},
  pages={10600--10611},
  year={2024},
  publisher={Royal Society of Chemistry}
}

@inproceedings{devlin2019bert,
  title={Bert: Pre-training of deep bidirectional transformers for language understanding},
  author={Devlin, Jacob and Chang, Ming-Wei and Lee, Kenton and Toutanova, Kristina},
  booktitle={Proceedings of the 2019 conference of the North American chapter of the association for computational linguistics: human language technologies, volume 1 (long and short papers)},
  pages={4171--4186},
  year={2019}
}

@incollection{lawson2014making,
  title={The making of reaxys—towards unobstructed access to relevant chemistry information},
  author={Lawson, Alexander J and Swienty-Busch, J{\"u}rgen and G{\'e}oui, Thibault and Evans, David},
  booktitle={The Future of the History of Chemical Information},
  pages={127--148},
  year={2014},
  publisher={ACS Publications}
}

@phdthesis{lowe2012extraction,
  title={Extraction of chemical structures and reactions from the literature},
  author={Lowe, Daniel Mark},
  year={2012}
}

@article{krallinger2015chemdner,
  title={The CHEMDNER corpus of chemicals and drugs and its annotation principles},
  author={Krallinger, Martin and Rabal, Obdulia and Leitner, Florian and Vazquez, Miguel and Salgado, David and Lu, Zhiyong and Leaman, Robert and Lu, Yanan and Ji, Donghong and Lowe, Daniel M and others},
  journal={Journal of cheminformatics},
  volume={7},
  number={Suppl 1},
  pages={S2},
  year={2015},
  publisher={Springer}
}

@article{jessop2011oscar4,
  title={OSCAR4: a flexible architecture for chemical text-mining},
  author={Jessop, David M and Adams, Sam E and Willighagen, Egon L and Hawizy, Lezan and Murray-Rust, Peter},
  journal={Journal of cheminformatics},
  volume={3},
  number={1},
  pages={41},
  year={2011},
  publisher={Springer}
}

@book{khabsa2015towards,
  title={Towards better accessibility of scholarly data},
  author={Khabsa, Madian},
  year={2015},
  publisher={The Pennsylvania State University}
}

@article{eltyeb2014chemical,
  title={Chemical named entities recognition: a review on approaches and applications},
  author={Eltyeb, Safaa and Salim, Naomie},
  journal={Journal of cheminformatics},
  volume={6},
  number={1},
  pages={17},
  year={2014},
  publisher={Springer}
}

@article{krallinger2017information,
  title={Information retrieval and text mining technologies for chemistry},
  author={Krallinger, Martin and Rabal, Obdulia and Lourenco, Analia and Oyarzabal, Julen and Valencia, Alfonso},
  journal={Chemical reviews},
  volume={117},
  number={12},
  pages={7673--7761},
  year={2017},
  publisher={ACS Publications}
}

@article{rocktaschel2012chemspot,
  title={ChemSpot: a hybrid system for chemical named entity recognition},
  author={Rockt{\"a}schel, Tim and Weidlich, Michael and Leser, Ulf},
  journal={Bioinformatics},
  volume={28},
  number={12},
  pages={1633--1640},
  year={2012},
  publisher={Oxford University Press}
}

@article{lample2016neural,
  title={Neural architectures for named entity recognition},
  author={Lample, Guillaume and Ballesteros, Miguel and Subramanian, Sandeep and Kawakami, Kazuya and Dyer, Chris},
  journal={arXiv preprint arXiv:1603.01360},
  year={2016}
}

@article{swain2016chemdataextractor,
  title={ChemDataExtractor: a toolkit for automated extraction of chemical information from the scientific literature},
  author={Swain, Matthew C and Cole, Jacqueline M},
  journal={Journal of chemical information and modeling},
  volume={56},
  number={10},
  pages={1894--1904},
  year={2016},
  publisher={ACS Publications}
}

@article{gupta2022matscibert,
  title={MatSciBERT: A materials domain language model for text mining and information extraction},
  author={Gupta, Tanishq and Zaki, Mohd and Krishnan, NM Anoop and Mausam},
  journal={npj Computational Materials},
  volume={8},
  number={1},
  pages={102},
  year={2022},
  publisher={Nature Publishing Group UK London}
}

@article{chithrananda2020chemberta,
  title={ChemBERTa: large-scale self-supervised pretraining for molecular property prediction},
  author={Chithrananda, Seyone and Grand, Gabriel and Ramsundar, Bharath},
  journal={arXiv preprint arXiv:2010.09885},
  year={2020}
}

@article{mallory2020extracting,
  title={Extracting chemical reactions from text using Snorkel},
  author={Mallory, Emily K and de Rochemonteix, Matthieu and Ratner, Alex and Acharya, Ambika and Re, Chris and Bright, Roselie A and Altman, Russ B},
  journal={BMC bioinformatics},
  volume={21},
  number={1},
  pages={217},
  year={2020},
  publisher={Springer}
}

@inproceedings{ratner2017snorkel,
  title={Snorkel: Rapid training data creation with weak supervision},
  author={Ratner, Alexander and Bach, Stephen H and Ehrenberg, Henry and Fries, Jason and Wu, Sen and R{\'e}, Christopher},
  booktitle={Proceedings of the VLDB endowment. International conference on very large data bases},
  volume={11},
  number={3},
  pages={269},
  year={2017}
}

@article{leaman2015tmchem,
  title={tmChem: a high performance approach for chemical named entity recognition and normalization},
  author={Leaman, Robert and Wei, Chih-Hsuan and Lu, Zhiyong},
  journal={Journal of cheminformatics},
  volume={7},
  number={Suppl 1},
  pages={S3},
  year={2015},
  publisher={Springer}
}

@inproceedings{dao2025entity,
  title={Entity-Based Synthetic Data Generation for Named Entity Recognition in Low-Resource Domains},
  author={Dao, Tuan An and Teranishi, Hiroki and Matsumoto, Yuji and Aizawa, Akiko},
  booktitle={JSAI International Symposium on Artificial Intelligence},
  pages={210--225},
  year={2025},
  organization={Springer}
}

@article{zhang2024chemical,
  title={A chemical reaction entity recognition method based on a natural language data augmentation strategy},
  author={Zhang, Xiaowen and Li, Yang and Li, Chaoyi and Zhu, Jingyuan and Gan, Zhiqiang and Wang, Lei and Sun, Xiaofei and You, Hengzhi},
  journal={Chemical Communications},
  volume={60},
  number={71},
  pages={9610--9613},
  year={2024},
  publisher={Royal Society of Chemistry}
}

@techreport{arthur2006k,
  title={k-means++: The advantages of careful seeding},
  author={Arthur, David and Vassilvitskii, Sergei},
  year={2006},
  institution={Stanford}
}

@article{houlsby2011bayesian,
  title={Bayesian active learning for classification and preference learning},
  author={Houlsby, Neil and Husz{\'a}r, Ferenc and Ghahramani, Zoubin and Lengyel, M{\'a}t{\'e}},
  journal={arXiv preprint arXiv:1112.5745},
  year={2011}
}

@inproceedings{gal2016dropout,
  title={Dropout as a bayesian approximation: Representing model uncertainty in deep learning},
  author={Gal, Yarin and Ghahramani, Zoubin},
  booktitle={international conference on machine learning},
  pages={1050--1059},
  year={2016},
  organization={PMLR}
}

@article{agrawal2021active,
  title={Active learning approach using a modified least confidence sampling strategy for named entity recognition},
  author={Agrawal, Ankit and Tripathi, Sarsij and Vardhan, Manu},
  journal={Progress in Artificial Intelligence},
  volume={10},
  number={2},
  pages={113--128},
  year={2021},
  publisher={Springer}
}

@article{liu2020ltp,
  title={LTP: a new active learning strategy for BERT-CRF based named entity recognition},
  author={Liu, Mingyi and Tu, Zhiying and Wang, Zhongjie and Xu, Xiaofei},
  journal={arXiv preprint arXiv:2001.02524},
  year={2020}
}

@inproceedings{scheffer2001active,
  title={Active hidden markov models for information extraction},
  author={Scheffer, Tobias and Decomain, Christian and Wrobel, Stefan},
  booktitle={International symposium on intelligent data analysis},
  pages={309--318},
  year={2001},
  organization={Springer}
}

@article{
kath2026the,
title={The Speed-up Factor: A Quantitative Multi-Iteration Active Learning Performance Metric},
author={Hannes Kath and Thiago S. Gouv{\^e}a and Daniel Sonntag},
journal={Transactions on Machine Learning Research},
issn={2835-8856},
year={2026},
url={https://openreview.net/forum?id=q6hRb6fETo},
note={}
}

@article{van2008visualizing,
  title={Visualizing data using t-SNE.},
  author={Van der Maaten, Laurens and Hinton, Geoffrey},
  journal={Journal of machine learning research},
  volume={9},
  number={11},
  year={2008}
}

@article{nachtegael2023study,
  title={A study of deep active learning methods to reduce labelling efforts in biomedical relation extraction},
  author={Nachtegael, Charlotte and De Stefani, Jacopo and Lenaerts, Tom},
  journal={PloS one},
  volume={18},
  number={12},
  pages={e0292356},
  year={2023},
  publisher={Public Library of Science San Francisco, CA USA}
}
\bibliographystyle{ieeetr}

\clearpage
\FloatBarrier

\appendix

\begin{table}[H]
\centering
\onecolumn
\caption{Test precision (\%) across rounds for product extraction. The best value in each strategy is in bold. The passive learning baseline is 84.62. }
\label{tab:prod_precision_rounds}
\begin{adjustbox}{max width=\textwidth}
\begin{tabular}{lcccccccccc}
\toprule
Strategy $\backslash$ Round & 1 & 2 & 3 & 4 & 5 & 6 & 7 & 8 & 9 & 10 \\
\midrule
Core-set & 0.00 & 66.67 & 74.65 & 77.22 & 72.92 & \textbf{81.82} & 73.50 & 73.79 & 72.82 & 77.11 \\
CLUSTER+ & 74.36 & 77.42 & 71.43 & 69.11 & 66.94 & 76.42 & 72.81 & 71.82 & 77.78 & \textbf{78.41} \\
BALD-batch & 72.16 & 80.33 & \textbf{84.38} & 77.78 & 82.81 & 81.94 & 82.61 & 76.39 & 81.32 & 77.23 \\
MLC & 70.24 & 73.61 & 69.86 & 75.38 & 80.00 & 82.61 & 80.56 & \textbf{88.46} & 85.06 & 73.47 \\
Margin & 52.14 & 72.00 & \textbf{94.55} & 86.15 & 87.10 & 85.71 & 81.71 & 77.91 & 77.32 & 76.04 \\
Entropy & 63.96 & 77.05 & 76.36 & \textbf{82.98} & 73.49 & 78.31 & 72.15 & 79.01 & 76.34 & 71.15 \\
Random & 73.91 & 73.47 & 68.85 & 67.27 & 70.09 & 72.82 & 73.96 & 77.08 & \textbf{77.65} & 76.40 \\
\bottomrule
\end{tabular}
\end{adjustbox}
\end{table}

\begin{table}[H]
\centering
\onecolumn
\caption{Test recall (\%) across rounds for product extraction. The best value in each strategy is in bold. The passive learning baseline is 69.37. }
\label{tab:prod_recall_rounds}
\begin{adjustbox}{max width=\textwidth}
\begin{tabular}{lcccccccccc}
\toprule
Strategy $\backslash$ Round & 1 & 2 & 3 & 4 & 5 & 6 & 7 & 8 & 9 & 10 \\
\midrule
Core-set & 0.00 & 3.60 & 47.75 & 54.95 & 63.06 & 64.86 & \textbf{77.48} & 68.47 & 67.57 & 57.66 \\
CLUSTER+ & 26.13 & 64.86 & \textbf{76.58} & 76.58 & 72.97 & 72.97 & 74.77 & 71.17 & 63.06 & 62.16 \\
BALD-batch & 63.06 & 44.14 & 48.65 & 44.14 & 47.75 & 53.15 & 51.35 & 49.55 & 66.67 & \textbf{70.27} \\
MLC & 53.15 & 47.75 & 45.95 & 44.14 & 54.05 & 51.35 & 52.25 & 62.16 & \textbf{66.67} & 64.86 \\
Margin & 65.77 & 48.65 & 46.85 & 50.45 & 48.65 & 59.46 & 60.36 & 60.36 & \textbf{67.57} & 65.77 \\
Entropy & 63.96 & 42.34 & 37.84 & 35.14 & 54.95 & 58.56 & 51.35 & 57.66 & 63.96 & \textbf{66.67} \\
Random & 61.26 & 64.86 & \textbf{75.68} & 66.67 & 67.57 & 67.57 & 63.96 & 66.67 & 59.46 & 61.26 \\
\bottomrule
\end{tabular}
\end{adjustbox}
\end{table}

\begin{table}[H]
\centering
\caption{Test precision (\%) across rounds for role labeling. The best value in each strategy is in bold. The passive learning baseline is 80.12. }
\label{tab:role_precision_rounds}
\begin{adjustbox}{max width=\textwidth}
\begin{tabular}{lcccccccccc}
\toprule
Strategy $\backslash$ Round & 1 & 2 & 3 & 4 & 5 & 6 & 7 & 8 & 9 & 10 \\
\midrule
Core-set & 0.00 & 52.97 & 64.35 & 68.63 & 71.27 & 77.81 & \textbf{79.69} & 77.40 & 78.30 & 78.26 \\
CLUSTER+ & 77.78 & 62.95 & 65.67 & 65.54 & 74.77 & 76.72 & 77.23 & 78.02 & 77.64 & \textbf{79.88} \\
BALD-batch & 63.25 & 59.58 & 66.38 & 69.69 & 68.61 & 73.39 & 74.63 & 73.82 & \textbf{76.60} & 75.68 \\
MLC & 0.00 & 64.41 & 67.11 & 71.01 & 73.65 & 75.22 & 77.98 & 77.88 & \textbf{78.25} & 76.81 \\
Margin & 45.45 & 59.92 & 64.55 & 66.86 & 74.55 & 75.38 & 76.62 & 75.38 & 76.56 & \textbf{77.81} \\
Entropy & 0.00 & 56.81 & 61.71 & 69.16 & 75.68 & 77.41 & 76.29 & 76.76 & 78.15 & \textbf{78.62} \\
Random & 0.00 & 73.26 & 59.79 & 64.64 & 68.62 & 66.85 & 70.36 & 71.79 & \textbf{74.71} & 72.83 \\
\bottomrule
\end{tabular}
\end{adjustbox}
\end{table}

\begin{table}[H]
\centering
\caption{Test recall (\%) across rounds for role labeling. The best value in each strategy is in bold. The passive learning baseline is 77.25.}
\label{tab:role_recall_rounds}
\begin{adjustbox}{max width=\textwidth}
\begin{tabular}{lcccccccccc}
\toprule
Strategy $\backslash$ Round & 1 & 2 & 3 & 4 & 5 & 6 & 7 & 8 & 9 & 10 \\
\midrule
Core-set & 0.00 & 55.99 & 69.16 & 73.35 & 75.75 & 74.55 & \textbf{77.54} & 74.85 & 74.55 & 75.45 \\
CLUSTER+ & 2.10 & 52.40 & 65.87 & 69.46 & 74.55 & 76.95 & 75.15 & 75.45 & 74.85 & \textbf{77.25} \\
BALD-batch & 22.16 & 59.58 & 68.56 & 73.65 & 73.95 & 75.15 & 74.85 & 75.15 & \textbf{75.45} & 74.55 \\
MLC & 0.00 & 42.81 & 61.08 & 71.86 & 73.65 & 75.45 & 76.35 & 76.95 & \textbf{77.54} & 76.35 \\
Margin & 1.50 & 47.01 & 63.77 & 68.86 & 73.65 & 74.25 & 74.55 & 74.25 & \textbf{77.25} & 74.55 \\
Entropy & 0.00 & 36.23 & 58.38 & 69.16 & 75.45 & \textbf{76.95} & 75.15 & 75.15 & 76.05 & 74.85 \\
Random & 0.00 & 18.86 & 50.30 & 66.77 & 70.06 & 71.26 & \textbf{76.05} & 75.45 & 76.95 & 75.45 \\
\bottomrule
\end{tabular}
\end{adjustbox}
\end{table}
\twocolumn
\end{document}